\documentclass[11pt]{article}

\usepackage{amsmath, amsthm, amscd, amsfonts, amssymb, graphicx, color}
\usepackage[bookmarksnumbered, colorlinks, plainpages]{hyperref}
\hypersetup{colorlinks=true,linkcolor=red, anchorcolor=green, citecolor=cyan, urlcolor=red, filecolor=magenta, pdftoolbar=true}
\usepackage{float}
\usepackage{multirow}
\usepackage{stackengine}
\usepackage{pdflscape}
\usepackage{tikz}
\usepackage{caption}
\usepackage{hyperref}
\usetikzlibrary{decorations.pathreplacing}
\usepackage{xcolor}
\usepackage{setspace}  % Package for line spacing
\usepackage{algorithm}
\usepackage{algpseudocode}
\usepackage{array}

\definecolor{BLUE}{rgb}{0,0,1}

\title{A Topology fixated Shape-Gradient Framework for Non-Simple Boundary Extraction for CIE-Lab  color images with Repulsive Energy}

\author{
	Shafeequdheen Palengara$^{1}$, Jyotiranjan Nayak$^{1}$, Vijayakrishna Rowthu$^{1}$\\
	\small $^{1}$ Tessellab, Department of Mathematics, SRM University-AP, India\\
	\small Emails: shafeequdheen\_p@srmap.edu.in, jyotiranjan\_n@srmap.edu.in, vijayakrishna.r@srmap.edu.in}

\begin{document}

\maketitle

Abstract
A level-set-free but a hybrid image segmentation approach based on a modified version of the piece-wise constant shape gradient of an Mumford–Shah shape functional and a repulsive function is considered. The segmentation is performed a non-local shape based through an evolution of discrete curves driven by a non-local shape based energy to segment images containing disjoint regions and multiple boundaries. This formulation has a novel additional component as a multivariable function dependent on a few sampled points of the curves that handles   the occurrence of self intersection during boundary curves evolution. The method is applied to a few gray-scale and color images, including images with nested structures and astronomical objects. The results indicate effective segmentation in complex scenarios with absolute control on the topology of the segments and self-intersections of the boundaries.\\



\section{Introduction}
Image segmentation is a fundamental problem in computer vision and image analysis, with important applications in object detection, medical imaging, and pattern recognition \cite{suri2000computer} \cite{gonzalez2009digital}\ \cite{zhang1996survey}. The goal of segmentation is to partition an image into meaningful regions based on characteristics such as intensity, texture, or color, thereby enabling effective interpretation and analysis. Among the various approaches to segmentation, variational methods have attracted sustained interest due to their solid mathematical foundation and their ability to handle noise, complex textures, and irregular boundaries \cite{mitiche2010variational} \cite{kass1988snakes} \cite{altinoklu2009image}.

A prominent variational model for image segmentation is the Mumford–Shah functional \cite{vitti2012mumford}, which formulates segmentation as an energy minimization problem by approximating an image with piecewise-smooth or piecewise-constant regions separated by a set of boundaries. Despite its theoretical appeal, direct minimization of the Mumford–Shah functional is challenging, and several approximations and numerical strategies have been proposed. One widely used simplification is the piecewise-constant model introduced by Chan and Vese, which employs level-set representations to evolve contours based on region statistics rather than explicit edge detection \cite{huang2021chan}\ \cite{liu2012local}\ \cite{brown2012completely} \cite{ruiying2022method}.

Level-set-based methods \cite{vese2002multiphase} \cite{brox2004level} \cite{tsai2001curve} \cite{han2003topology} provide flexibility in handling topological changes during boundary evolution; however, they also introduce additional computational complexity due to reinitialization procedures, numerical stability constraints, and time-step restrictions. Moreover, automatic topology changes are not always desirable, particularly in applications where the preservation of boundary structure or prior knowledge about object geometry is important.

An alternative framework for variational segmentation is provided by shape calculus \cite{delfour2011shapes} \cite{sokolowski1992introduction}, in which segment regions are treated as geometric objects whose evolution is governed by the shape derivative of an energy functional \cite{walker2015shapes}. In this setting, the boundary is evolved directly through a velocity field derived from the shape gradient, avoiding the need for implicit level-set representations. Shape-gradient-based methods \cite{jehan2006shape}\cite{tsai2003shape}, \cite{aubert2003image} offer explicit boundary control without edge based information, reduced computational and memory  overhead, and a clearer geometric interpretation of boundary evolution.

Within this framework, a level-set-free segmentation method based on the shape gradient of the modified  piecewise-constant Mumford–Shah energy, which we call it as Normalized Mumford-Shah Energy (NMS) here onwards was previously introduced \cite{nayak2025shape}, where the evolving boundary was simple in topology. The sampled vertices of the boundary  which resemble a polygon were updated iteratively using a gradient descent scheme driven by the shape derivative \cite{walker2015shapes}\ \cite{sokolowski1992introduction}. This approach demonstrated effective segmentation for gray-scale and color images, particularly in capturing intricate and highly concave boundaries, while maintaining computational efficiency.

Despite these advantages, the single-boundary(simple curve) representation presents practical  limitations when dealing with images containing multiple disjoint objects, nested boundaries, or complex scenes requiring separate boundaries. In addition, the sample points of the boundary evolution may lead to self-intersections when vertices or edges move too close during deformation, potentially resulting in unstable evolution and incorrect segmentation outcomes in some cases.

To overcome these limitations, the present work extends the shape-gradient-based segmentation framework in two directions. First, multiple boundary curves are introduced to enable the simultaneous evolution of independent boundaries within a unified variational setting. This extension allows the method to handle multi-object segmentation, disconnected components, and images with multiple boundaries. Second, a repulsive interaction energy  term is added to the shape energy functional to discourage proximity between boundary edges, thereby reducing self-intersection during evolution and preserving valid boundary topology. This resulting in a hybrid form of energy between shape functional and a multivariable function.

The resulting hybrid framework retains the advantages of explicit, levelset-free boundary  representation while providing improved robustness and flexibility in complex segmentation scenarios by strict restriction on topology. Experimental results on gray-scale and color images demonstrate that the proposed approach effectively segments images containing multiple objects and closely spaced boundaries, while maintaining stable and non-self-intersecting boundary evolution.

\section{Modified Mumford--Shah Shape Model with Repulsive Energy for Multiple Boundaries}

\subsection{Multiple-Boundary Representation}

\begin{figure}[H]
    \centering
    \begin{minipage}{0.48\linewidth}
        \centering
        \includegraphics[width=\linewidth]{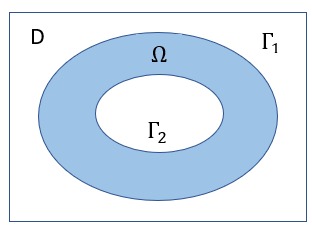}
        \caption*{(a)}
    \end{minipage}
    \hfill
    \begin{minipage}{0.48\linewidth}
        \centering
        \includegraphics[width=\linewidth]{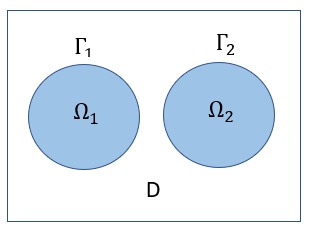}
        \caption*{(b)}
    \end{minipage}
    \caption{Multiple-boundary configurations:
    (a) a single region with a hole, and
    (b) two disjoint regions inside the domain \(D\).}
    \label{fig:multi-configuration}
\end{figure}

In many image segmentation problems \cite{chan2001active} \cite{caselles1997geodesic}\ \cite{li2010distance}, objects may possess more than one boundary. 
Such boundaries may either correspond to a single object containing holes or to 
several disjoint objects. Figure~\ref{fig:multi-configuration} illustrates these 
two situations. In Figure~\ref{fig:multi-configuration}(a), a single object 
occupies the region \(\Omega\) and contains a hole. Consequently, the object is 
bounded by two closed curves: an outer boundary \(\Gamma_1\) and an inner boundary 
\(\Gamma_2\). In contrast, Figure~\ref{fig:multi-configuration}(b) shows two 
disjoint objects, each enclosed by its own boundary \(\Gamma_1\) and 
\(\Gamma_2\), defining separate regions \(\Omega_1\) and \(\Omega_2\).

In general, the segmented region \(\Omega\) and its boundary \(\Gamma\) are defined as
\[
\Omega = \bigcup_{k=1}^{N_{1}} \Omega_k,
\qquad
\Gamma = \bigcup_{k=1}^{N_{2}} \Gamma_k,
\]
where each closed curve \(\Gamma_k\) encloses the corresponding region 
\(\Omega_k\). The complement of the segmented regions is denoted by \(\Omega^c\).

Modified piecewise-constant Mumford--Shah shape functional(mpMS) for multiple boundaries is
\[
E(\Gamma)
=
\sum_{k=1}^{N_1}
\frac{\alpha}{|\Omega_k|}
\int_{\Omega_k} \big(f(x)-\mu(\Omega_k)\big)^2 \, dx
+
\frac{\beta}{|\Omega^c|}
\int_{\Omega^c} \big(f(x)-\mu(\Omega^c)\big)^2 \, dx
+
\eta \sum_{k=1}^{N_2} \int_{\Gamma} d\Gamma ,
\]
where \(f\) denotes the image intensity, \(\mu(\cdot)\) denotes region means, and 
\(\eta>0\) controls boundary regularization.

\subsection{Motivation for Repulsive Energy}

A major challenge in variational image segmentation with a sampled discrete curve 
representations is the occurrence of self-intersections, particularly when 
multiple discrete curves evolve simultaneously. Such intersections lead to 
topologically invalid boundaries, numerical instability, and inaccurate 
segmentation.

While level set methods \cite{osher2001level}, \cite{gibou2018review} naturally handle topology 
changes, shape-based methods operating directly on sampled curves offer 
interpretability and computational efficiency. However, without explicit 
constraints, gradient-based vertex updates may result in intersecting or degenerate boundaries.

To address this issue, a repulsive energy term is incorporated into the 
Mumford--Shah framework. This term penalizes geometric configurations in which 
the boundary edges intersect or become  collinear but close, thereby preserving valid topology 
during contour evolution.

\subsection{Mathematical Formulation of Repulsive Energy}

Consider two line segments $S_1$ and $S_2$ defined by the points $A(x_1,y_1)$, $B(x_2,y_2)$ and $A'(x_3,y_3)$, $B'(x_4,y_4)$, respectively, defined by
\begin{align}
  S_1 &= A + \lambda\,(B-A), \qquad \lambda\in[0,1],\\[3pt]
  S_2 &= A' + \mu\,(B'-A'), \qquad \mu\in[0,1].
\end{align}

We distinguish different cases for the interaction of these segments:
\begin{itemize}
  \item $S_1$ and $S_2$ intersect at a single point.
  \item $S_1$ and $S_2$ intersect at infinitely many points.
  \item $S_1$ and $S_2$ are parallel.
\end{itemize}
There are different cases for the intersection of line segments: intersection at a unique point and intersection at infinitely many points. We have formulated a repulsive energy function for each case.
\subsection*{Energy function to avoid unique intersection}
\begin{figure}[H]
  \centering
  \begin{minipage}{0.23\linewidth}\centering
    \includegraphics[width=\linewidth]{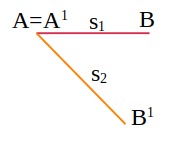}\\[3pt]$\lambda=\mu=0$
  \end{minipage}\hfill
  \begin{minipage}{0.23\linewidth}\centering
    \includegraphics[width=\linewidth]{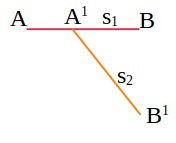}\\[3pt]$\lambda\in(0,1),\;\mu=0$
  \end{minipage}\hfill
  \begin{minipage}{0.23\linewidth}\centering
    \includegraphics[width=\linewidth]{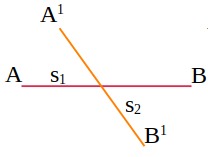}\\[3pt]$\lambda,\mu\in(0,1)$
  \end{minipage}\hfill
  \begin{minipage}{0.23\linewidth}\centering
    \includegraphics[width=\linewidth]{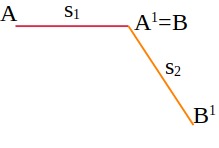}\\[3pt]$\lambda=1,\;\mu=0$
  \end{minipage}
  \caption{Four possible intersections of two line segments for different values of $\lambda$ and $\mu$.}
  \label{fig:unique_cases}
\end{figure}

If $S_1$ and $S_2$ intersect exactly once, then there exist unique parameters $\lambda,\mu\in[0,1]$ (see Fig.~\ref{fig:unique_cases}) satisfying
\begin{equation}\label{eq:system}
  A + \lambda\,(B-A) \;=\; A' + \mu\,(B'-A').
\end{equation}

\begin{figure}[H]
  \centering
  \begin{minipage}{0.48\linewidth}\centering
    \includegraphics[width=\linewidth]{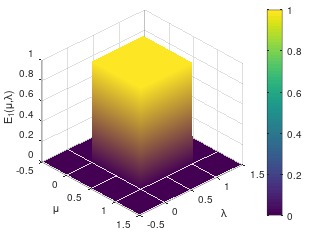}\\[6pt]
    Indicator function $I(\mu,\lambda)$.
  \end{minipage}\hfill
  \begin{minipage}{0.46\linewidth}\centering
    \includegraphics[width=\linewidth]{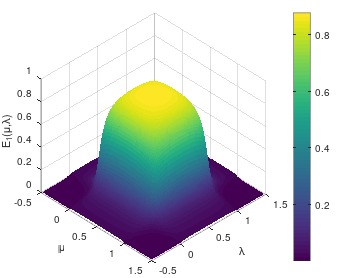}\\[6pt]
    Smooth approximation $E_1(\mu,\lambda)$.
  \end{minipage}
  \caption{Comparison of the indicator function $I(\mu,\lambda)$ (left) with its smooth approximation $E(\mu,\lambda)$ (right) with \(\varepsilon=0.05\).}
  \label{fig:indicator_vs_approx1}
\end{figure}

Equation~\eqref{eq:system} can be rewritten as
\begin{align}
  A-A' &= \mu\,(B'-A') + \lambda\,(A-B)\\[3pt]
        &= \begin{bmatrix} B'-A' & A-B \end{bmatrix}
           \begin{bmatrix} \mu \\ \lambda \end{bmatrix}.
\end{align}
\noindent Consequently,
\begin{align}
  \begin{bmatrix} x_1-x_3 \\ y_1-y_3 \end{bmatrix}
  &=
  \begin{bmatrix}
    x_4-x_3 & x_1-x_2\\
    y_4-y_3 & y_1-y_2
  \end{bmatrix}
  \begin{bmatrix} \mu \\ \lambda \end{bmatrix}.
\end{align}
Defining the determinant
\begin{equation}
\Delta\;=\;(x_4-x_3)(y_1-y_2)-(x_1-x_2)(y_4-y_3),\label{determinat}
\end{equation}
we obtain
\begin{align}
  \begin{bmatrix} \mu \\ \lambda \end{bmatrix}
  &=\frac{1}{\Delta}
    \begin{bmatrix}
      y_1-y_2 & x_2-x_1\\
      y_3-y_4 & x_4-x_3
    \end{bmatrix}
    \begin{bmatrix} x_1-x_3 \\ y_1-y_3 \end{bmatrix}.
\end{align}
The repulsive energy is activated only when $0\le\mu,\lambda\le1$.  We therefore define the indicator function
\begin{equation}
  \mathbb{I}(\mu,\lambda)=\begin{cases}
    1,& 0\le\mu,\lambda\le1,\\[3pt]
    0,& \text{otherwise}.
  \end{cases}
\end{equation}
The Energy \(E_s\) defined as smooth approximation of $\mathbb{I}$ is given by (see Fig.~\ref{fig:indicator_vs_approx1})
\begin{equation}
E_{s}(\mu,\lambda)=\frac{1}{\pi^2}\Bigl(\tan^{-1}\Bigl(\frac{\mu}{\varepsilon}\Bigl)-\tan^{-1}\Bigl(\frac{\mu-1}{\varepsilon}\Bigr)\Bigr)
                  \Bigl(\tan^{-1}\Bigl(\frac{\lambda}{\varepsilon}\Bigr)-\tan^{-1}\Bigl(\frac{\lambda-1}{\varepsilon}\Bigr)\Bigr).
\end{equation}

\subsection*{Angle-based energy to repel parallel and close segments.}

Another serious issue with point-wise updates is mutual twisting of two adjacent curve segments onto each other. To discourage two line segments to be parallel particularly when the segments are close enough, we design an angle-based energy function between two adjacent line segments \(S_1\) and \(S_2\).  
If the angle between the two segments is \(0^\circ\) or \(180^\circ\), then their extensions overlap and the line segments share infinitely many intersection points. This degenerate configuration must be avoided before the onset. Therefore, an energy function is introduced that depends on the angle between adjacent edges at each vertex. The energy is controlled through the inner product of two adjacent vectors at every vertex.

Consider two vectors adjacent to the vertex \(V_i\):
\[
\vec{S}_{i-1} = V_{i-1} - V_i, 
\qquad 
\vec{S}_{i} = V_{i+1} - V_i.
\]

Let \(D_i = \vec{S}_{i-1} \cdot \vec{S}_{i}\) denote their dot product. Since
\[
D_i = \|\vec{S}_{i-1}\|\,\|\vec{S}_{i}\| \cos\theta_i,
\]
the dot product approaches its extreme values when \(\theta_i \to 0^\circ\) or \(\theta_i \to 180^\circ\), that is, when the two segments become parallel. In order to softly penalize this configuration, a smooth barrier-type energy is defined as
\[
E_{\theta_{i}}
= \frac{1}{\pi}
\left[
\tan^{-1}\!\left(\frac{D_i + \sigma}{\varepsilon}\right)
-
\tan^{-1}\!\left(\frac{D_i - \sigma}{\varepsilon}\right)
\right],
\]
where \(\sigma > 0\) controls the width of the critical region around the collinear configuration, and \(\varepsilon > 0\) is a small regularization parameter that ensures smoothness and numerical stability.

This formulation behaves as a smooth indicator function and about to  becomes active when \( |D_i| < \sigma \), that is, when the two line segments are close and about to being collinear. When the configuration is safe, the energy remains almost constant and does not affect the system. Since the function is smooth and differentiable everywhere, it is well suited for gradient-based optimization methods.

\begin{figure}[H]
  \centering
  \begin{minipage}{0.47\linewidth}\centering
    \includegraphics[width=\linewidth]{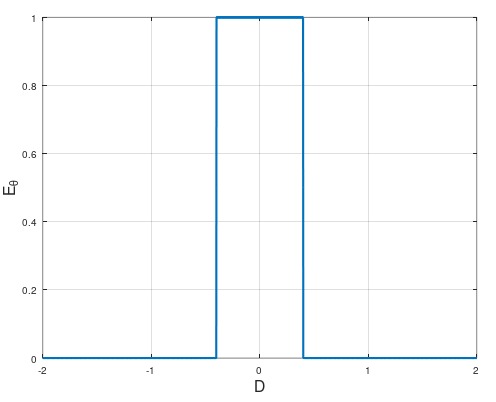}\\[6pt]
    Indicator function $E_{\theta}(\sigma=0.4,\epsilon=10^{-5})$.
  \end{minipage}\hfill
  \begin{minipage}{0.47\linewidth}\centering
    \includegraphics[width=\linewidth]{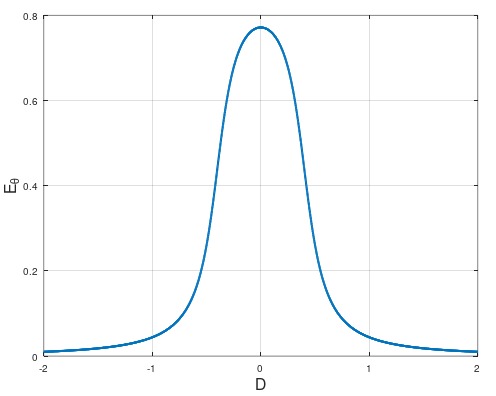}\\[6pt]
    Smooth approximation $E_{\theta}(\sigma=0.4,\epsilon=0.15)$.
  \end{minipage}
  \caption{Comparison of the indicator function $E_{\theta}((\sigma=0.4,\epsilon=10^{-5})$ (left) with its smooth approximation $E_{\theta} (\sigma=0.4,\epsilon=0.15)$  with \(\varepsilon=0.05\)\ (right).}
  \label{fig:indicator_vs_approx}
\end{figure}

\subsection*{Repulsive Energy}
Using these two energy functions, we formulate a repulsive energy \(E_{R}\) to prevent the intersection of any two line segments \(S_1\) and \(S_2\).

\begin{equation}
E_R = \left(\frac{1}{\varepsilon} - \delta(\Delta)\right) E_s
      + \delta(\Delta)\, E_{\theta}
\end{equation}
Here, \(\Delta\) is the determinant defined in Equation \ref{determinat}, and \(\delta(\cdot)\) is the Dirac delta function.
When \(\Delta \neq 0\), it corresponds to the unique intersection case, for which $\delta(\Delta) \simeq 0$,
and the energy \(E_s\) is activated otherwise \(E_\theta\) will be active.

\subsection{Working Principle of Repulsive Energy in Image Segmentation}

\begin{figure}[h]
  \centering
  \includegraphics[width=0.5\textwidth]{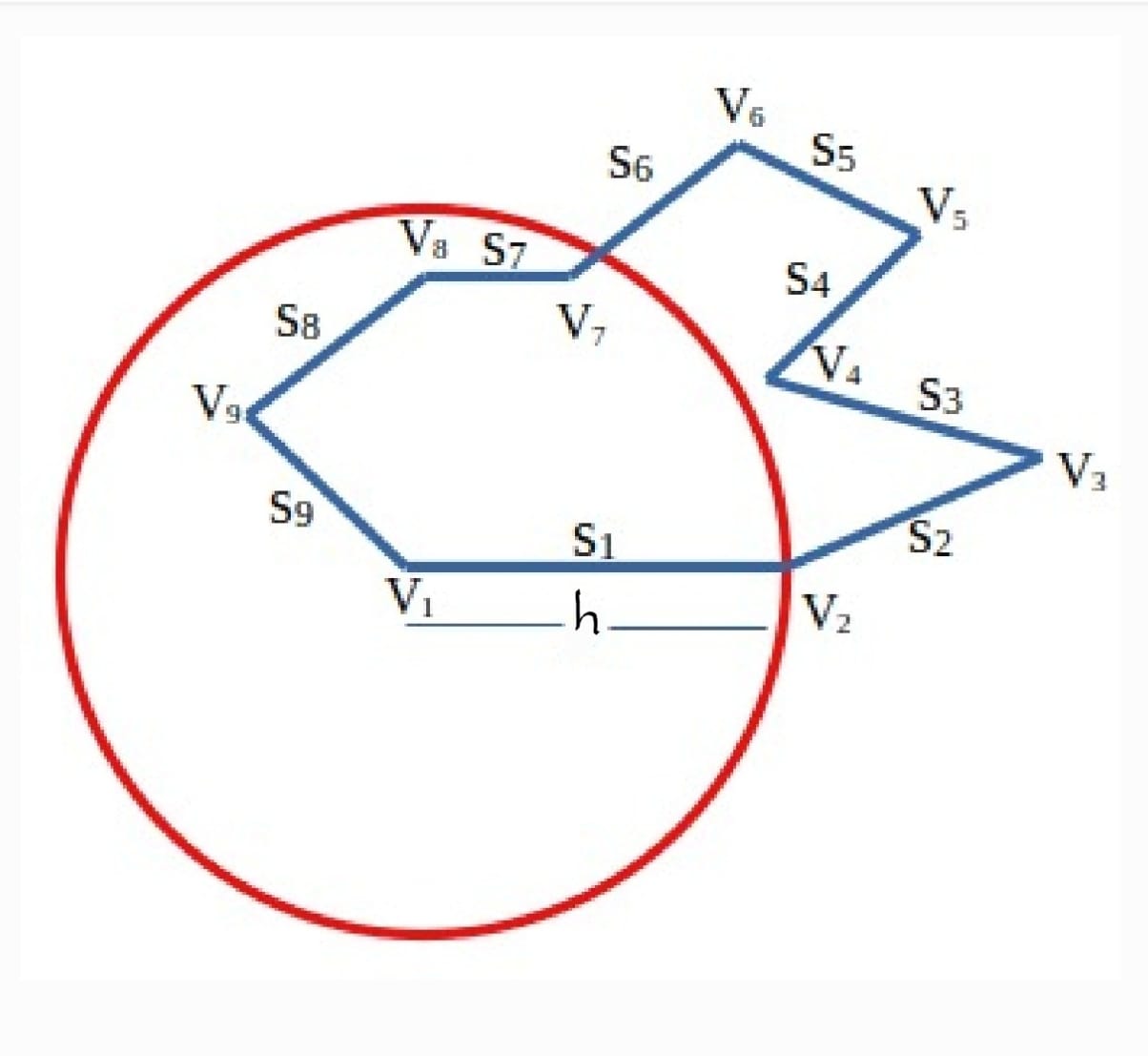}
  \caption{Illustration of $h$-nbd around vertex $v_1$ and its surrounding edges. Adjacent edges are excluded from repulsion computation.}
  \label{fig:repulsive-vertex}
\end{figure}

For a given vertex $v_i$, repulsive
interactions are computed between edges incident to $v_i$ and other nearby edges
that are \emph{not directly connected} to it. As illustrated in
Fig.~\ref{fig:repulsive-vertex}, these interactions are classified into two
geometric cases, corresponding to unique intersections and collinear (infinite)
intersections.

Let $s_i=(v_{i-1},v_i)$ denote an edge adjacent to $v_i$, and let
$s_j=(v_{j-1},v_j)$ be any other edge on the boundary. Repulsive energy
is defined only for \emph{non-adjacent} edge pairs, since adjacent edges share a
common vertex and cannot intersect in a meaningful geometric sense.
Consequently, all edges directly connected to $v_i$, as well as their immediate
neighbors, are excluded from the repulsion computation.

To illustrate this process, consider the evaluation of repulsive energy at vertex
$v_1$ in Fig.~\ref{fig:repulsive-vertex}. The edges incident to $v_1$ are $s_1$
and $s_9$. Since $s_1$ is adjacent to $s_2$ and $s_9$, and $s_9$ is adjacent to
$s_8$, all these edges within the neighbourhood are excluded from the repulsion evaluation to avoidredundant or invalid interactions, remaining with  s7 only to compute the intersection part of the repulsion.

\paragraph{Selective Evaluation of $E_R$ in $h$-Neighbourhood.}
To reduce computational complexity, repulsive interactions are evaluated only
within a localized region around each vertex. The $h$-neighbourhood
($h$-nbd) of a vertex $v_i$ is defined as a circular region centered at
$v_i$ with radius $h_i$, where $h_i$ is chosen as the maximum length of
the edges incident to $v_i$. All vertices lying inside this region are
identified, and the edges connected to those vertices are collected as potential
candidates for repulsive interaction.

In Fig.~\ref{fig:repulsive-vertex}, the $h$-nbd of
$v_1$ contains the vertices $v_{10}$, $v_9$, and $v_7$. Accordingly, the
associated edges $s_1$, $s_2$, $s_6$, $s_7$, $s_8$, and $s_9$ are initially
selected. From this set, edges adjacent to $v_1$ (namely $s_1$ and $s_9$) and
their immediate neighbors ($s_2$ and $s_8$) are excluded, leaving only
geometrically valid, non-adjacent edges for repulsion computation. Consequently,
the valid edge pairs contributing to the repulsive energy at vertex $v_1$ are
\[
(s_1, s_7), \quad (s_1, s_6), \quad (s_9, s_7), \quad (s_9, s_6).
\]
The repulsive energy corresponding to these edge pairs is then evaluated using
the formulations introduced in the previous section.

This localized and selective evaluation strategy significantly reduces the
number of edge comparisons while preserving the effectiveness of the repulsive
force. As a result, the evolving polygonal contours remain free from
self-intersections and degeneracies, enabling stable and accurate segmentation
in complex multi-boundary configurations.

\section{Gradient of the Modified Mumford--Shah Energy and Repulsive Energy}
\label{Shape Derivative Modified Mumford and Shah Model}

For the proposed multi-boundary segmentation framework, the total energy
functional is decomposed into four components as
\begin{align}\label{Energy_components_new}
E &= \sum_{k=1}^{N} E_{1}(\Omega_k) + E_{2}(\Omega^c) + E_{3}(\Gamma) + \lambda E_R(\Gamma),
\end{align}
where $\Omega_k$ denotes the $k$-th segmented region with boundary $\Gamma_k$,
$\Omega^c$ is the background region, and $\Gamma=\bigcup_{k=1}^{N}\Gamma_k$.

The region-based energy terms are defined as
\begin{align}
E_{1}(\Omega_k) &= \frac{1}{|\Omega_k|}
\int_{\Omega_k} \big(f(x)-\mu(\Omega_k)\big)^2 \, \mathrm{d}x, \\
E_{2}(\Omega^c) &= \frac{1}{|\Omega^c|}
\int_{\Omega^c} \big(f(x)-\mu(\Omega^c)\big)^2 \, \mathrm{d}x,
\end{align}
and the boundary regularization term is
\begin{equation}
E_{3}(\Gamma)=\sum_{k=1}^{N}\int_{\Gamma_k}1\ \mathrm{d}\Gamma .
\end{equation}

Substituting the expressions for the region means into $E_{1}(\Omega_k)$ yields
\begin{align}
E_{1}(\Omega_k)
=
\frac{1}{|\Omega_k|}
\left(
\int_{\Omega_k} f(x)^2\,\mathrm{d}x
-
\frac{1}{|\Omega_k|}
\left(\int_{\Omega_k} f(x)\,\mathrm{d}x\right)^2
\right).
\end{align}
The corresponding shape derivative is given by,
\begin{align}
\delta E_{1}(\Omega_k)
&=
\frac{1}{|\Omega_k|} f(x)^2
-
\frac{1}{|\Omega_k|^2}
\left(
\int_{\Omega_k} f(x)^2\,\mathrm{d}x
+
2 f(x)\int_{\Omega_k} f(x)\,\mathrm{d}x
-
\frac{2}{|\Omega_k|}
\left(\int_{\Omega_k} f(x)\,\mathrm{d}x\right)^2
\right).
\end{align}

Since $\Omega^c = D \setminus \bigcup_{k=1}^{N}\Omega_k$, the shape derivative of
$E_{2}(\Omega^c)$ is obtained analogously and is given by
\begin{align}
\delta E_{2}(\Omega^c)
&=
-\frac{1}{|\Omega^c|} f(x)^2
+
\frac{1}{|\Omega^c|^2}
\left(
\int_{\Omega^c} f(x)^2\,\mathrm{d}x
-
\frac{2}{|\Omega^c|}
\left(\int_{\Omega^c} f(x)\,\mathrm{d}x\right)^2
\right).
\label{shape_gradient_E2_new}
\end{align}

The boundary regularization term introduces curvature-dependent motion. Using
standard results from shape calculus \cite{sokolowski1992introduction}, its shape derivative is given by
\begin{align}
\delta E_{3}(\Gamma)
&=
\sum_{k=1}^{N}\int_{\Gamma_k} \kappa_k \, \mathrm{d}\Gamma ,
\label{gradient_of_E3_new}
\end{align}
where $\kappa_k$ denotes the curvature of the boundary $\Gamma_k$.

Combining all contributions, the shape derivative of the shape functional part is given by
\begin{equation}
\delta E
=
\sum_{k=1}^{N} \delta E_{1}(\Omega_k)
+
\delta E_{2}(\Omega^c)
+
\delta E_{3}(\Gamma).
\end{equation}

The boundary evolution is driven in the normal direction $\hat{n}$ according to a
gradient descent scheme, yielding a stable evolution of multiple interacting
polygonal curves.

\subsection{Gradient of the Repulsive Energy}
In the proposed model, the repulsive energy is formulated to distinguish between different geometric configurations of line segments, namely uniquely intersecting and collinearly intersecting cases. The first component of the energy functional, \(E_s\), helps to prevent the unique intersection of two line segments, while the second term, \(E_{\theta}\), helps to prevent coincident line segments. 

To enable effective optimization, it is essential to compute the gradient of the total repulsive energy with respect to the vertex coordinates. This gradient captures the sensitivity of the energy to infinitesimal perturbations in the vertex positions and plays a central role in vertex-based energy minimization. In this section, we derive the gradients for each individual energy term and then present the complete gradient of the total repulsive energy.
\subsection{Gradient of \(E_{s}\)}
The energy \(E_{s}(\mu,\lambda)\) can be written as
\begin{equation}
E_{s}(\mu,\lambda)=F(\lambda)F(\mu)
\end{equation}
Here, \( F(x)\) defined by
\begin{equation}
F(x) = tan^{-1}(x) -tan^{-1}(x-1)
\end{equation}
The parameters $\mu$ and $\lambda$ are given by
\begin{align}
  \mu &= \frac{(y_1-y_2)(x_1-x_3)+(x_2-x_1)(y_1-y_3)}{(y_1-y_2)(x_4-x_3)-(x_2-x_1)(y_3-y_4)},\\[6pt]
  \lambda &= \frac{(y_3-y_4)(x_1-x_3)+(x_4-x_3)(y_1-y_3)}{(y_1-y_2)(x_4-x_3)-(x_2-x_1)(y_3-y_4)}.
\end{align}
Define the numerators
\begin{equation}
  N_1 = (y_1-y_2)(x_3-x_4)+(x_2-x_1)(y_1-y_3),\\
  N_2 = (y_3-y_4)(x_1-x_3)+(x_4-x_3)(y_1-y_3),
\end{equation}
with common denominator
\begin{equation}
  \Delta = (y_1-y_2)(x_4-x_3)-(x_2-x_1)(y_3-y_4).
\end{equation}
The partial derivatives of $\mu$ read
\begin{align}
  \frac{\partial \mu}{\partial x_1} &= \frac{\Delta(y_3-y_2)-N_1(y_3-y_4)}{\Delta^2},\\[3pt]
  \frac{\partial \mu}{\partial y_1} &= \frac{\Delta(x_2-x_3)-N_1(x_4-x_3)}{\Delta^2},
\end{align}
while those of $\lambda$ are
\begin{align}
  \frac{\partial \lambda}{\partial x_1} &= \frac{(\Delta-N_2)(y_3-y_4)}{\Delta^2},\\[3pt]
  \frac{\partial \lambda}{\partial y_1} &= \frac{(\Delta-N_2)(x_4-x_3)}{\Delta^2}.
\end{align}
Consequently,
\begin{align}
  \frac{\partial E_s}{\partial x_1}
  &= F(\mu)\left(\frac{1}{1+\lambda^2}-\frac{1}{1+(1-\lambda)^2}\right)\frac{\partial \lambda}{\partial x_1}
     +F(\lambda)\left(\frac{1}{1+\mu^2}-\frac{1}{1+(1-\mu)^2}\right)\frac{\partial \mu}{\partial x_1},\\[6pt]
  \frac{\partial E_s}{\partial y_1}
  &= F(\mu)\left(\frac{1}{1+\lambda^2}-\frac{1}{1+(1-\lambda)^2}\right)\frac{\partial \lambda}{\partial y_1}
     +F(\lambda)\left(\frac{1}{1+\mu^2}-\frac{1}{1+(1-\mu)^2}\right)\frac{\partial \mu}{\partial y_1}.
\end{align}

\subsection{Gradient of \(E_{\theta}\)}
Consider the angle-based energy \(E_{\theta i}\) defined at each vertex \(v_i\), with two segments adjacent to the vertex \(v_i\), namely \(S_i\) and \(S_{i-1}\).

\[
\vec{S}_i = (x_{i+1}-x_i,\; y_{i+1}-y_i)
\]

\[
\vec{S}_{i-1} = (x_{i-1}-x_i,\; y_{i-1}-y_i)
\]

The dot product \(\vec{S}_{i-1} \cdot \vec{S}_i\) is denoted as \(D_i\) given by

\begin{align*}
D_i
&= (x_{i+1} - x_i)(x_{i-1} - x_i)
+ (y_{i+1} - y_i)(y_{i-1} - y_i) \\
&= x_{i-1}x_{i+1} - x_{i-1}x_i - x_ix_{i+1} + x_i^2 + y_{i-1}y_{i+1} - y_{i-1}y_i - y_iy_{i+1} + y_i^2
\end{align*}

Then the partial derivatives are

\[
\frac{\partial D_i}{\partial x_i} = -x_{i-1} - x_{i+1} + 2x_i
\]

\[
\frac{\partial D_i}{\partial y_i} = -y_{i-1} - y_{i+1} + 2y_i
\]

Therefore, the gradient of $E_\theta$ is defined as
\[
\nabla E_\theta = (dE_{\theta x},\; dE_{\theta y})
\]

where

\[
dE_{\theta x}
= \frac{1}{1 + D_i^2}\,(2x_i - x_{i-1} - x_{i+1})
\]

\[
dE_{\theta y}
= \frac{1}{1 + D_i^2}\,(2y_i - y_{i-1} - y_{i+1})
\]

\section{Experimental Validation}\label{Experimental validation}

A series of numerical experiments were conducted to evaluate the effectiveness
of the proposed repulsive energy in preventing self-intersections and self-correction  the
evolution of boundary segments under different initialization scenarios. In
the proposed framework, two distinct repulsive mechanisms act on each boundary
vertex. The first repulsive term, denoted by $E_s$, is designed to prevent
non-parallel intersections between boundary edges, while the
second term, $E_{\theta}$, addresses degenerate configurations arising from
nearly parallel or collinear edges that may intersect at infinitely many points.

\begin{figure}[H]
    \centering
    % First Row
    \begin{minipage}{0.33\linewidth}
        \centering
        \includegraphics[width=\linewidth]{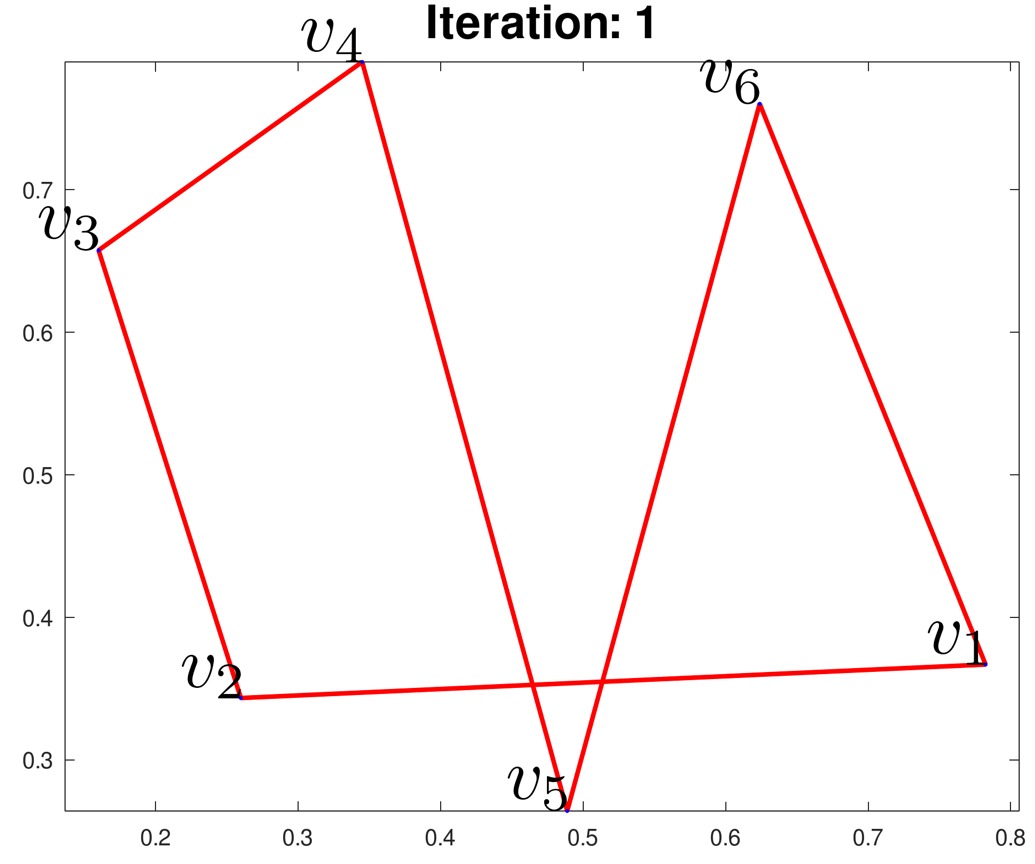}
    \end{minipage}
    \begin{minipage}{0.33\linewidth}
        \centering
        \includegraphics[width=\linewidth]{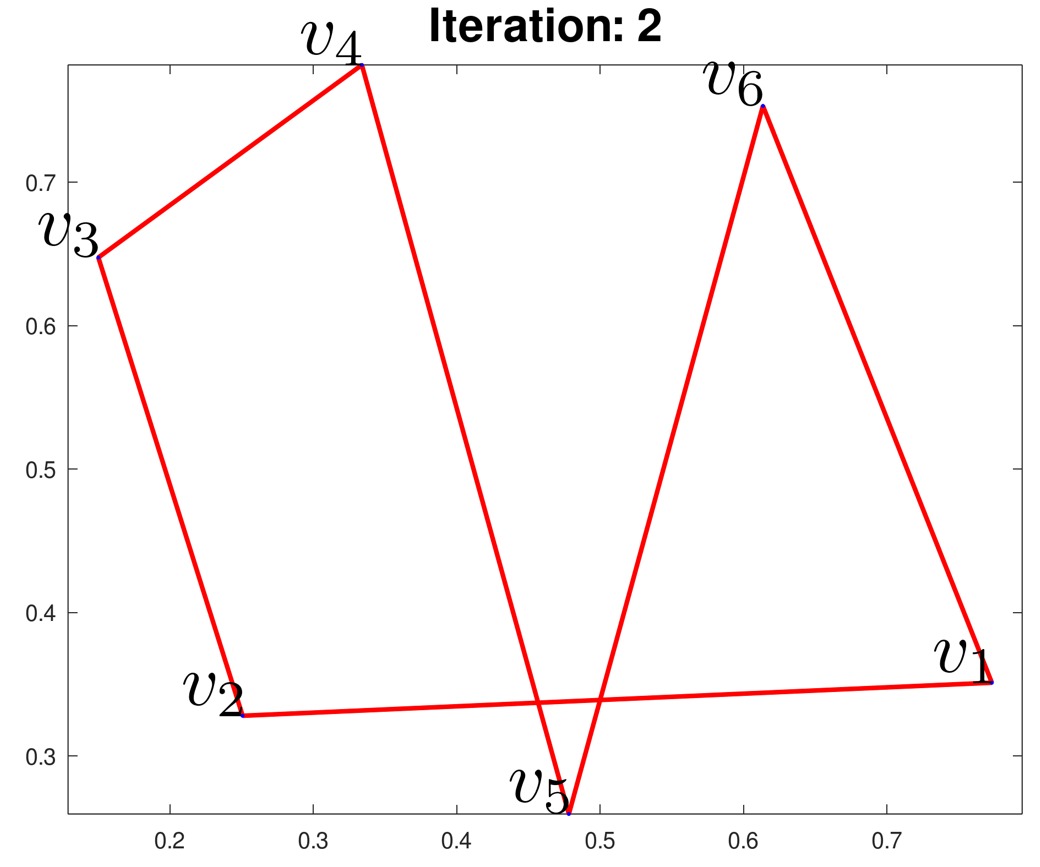}
    \end{minipage} 
    \begin{minipage}{0.32\linewidth}
        \centering
        \includegraphics[width=\linewidth]{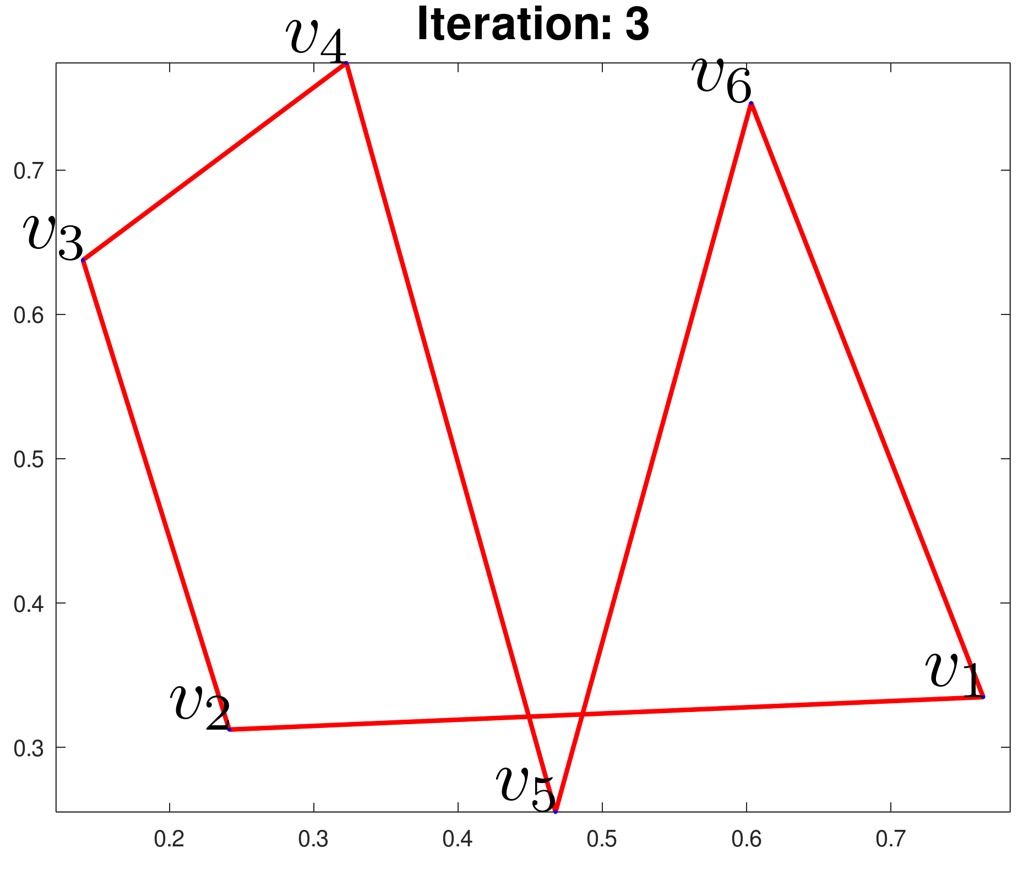}
    \end{minipage}  \\[0.5em]

    \begin{minipage}{0.33\linewidth}
        \centering
        \includegraphics[width=\linewidth]{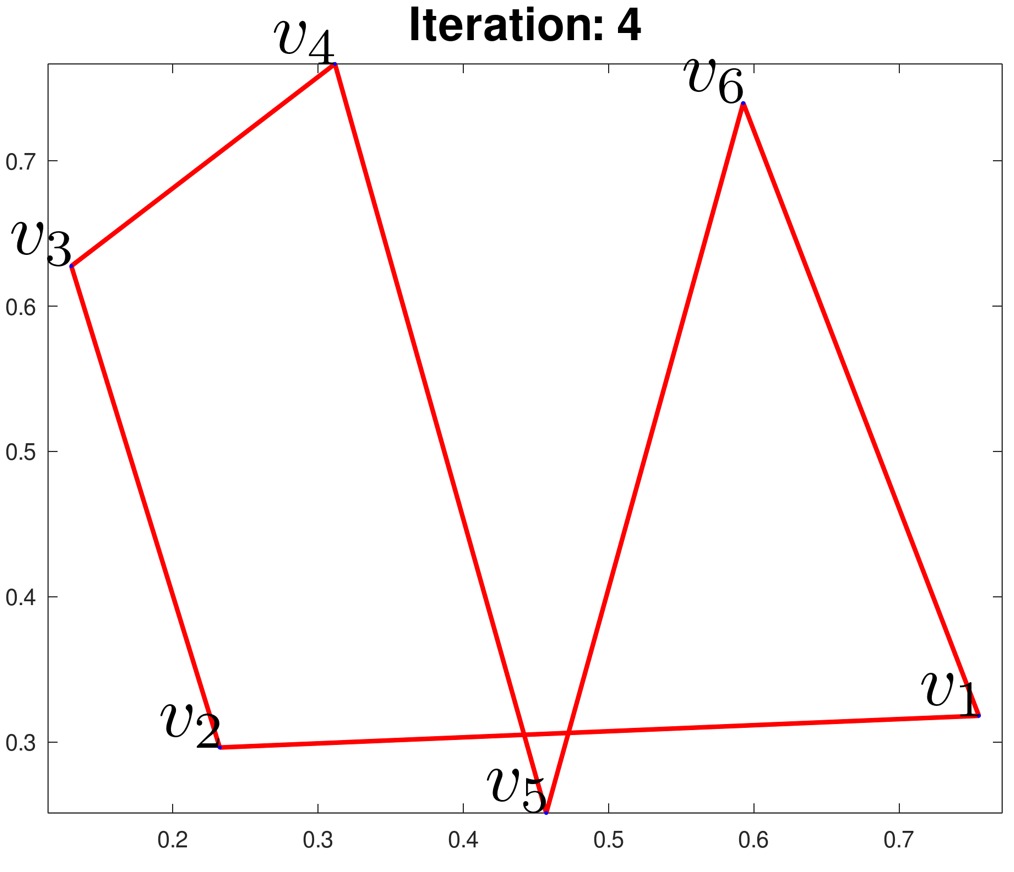}
    \end{minipage}
    \begin{minipage}{0.33\linewidth}
        \centering
        \includegraphics[width=\linewidth]{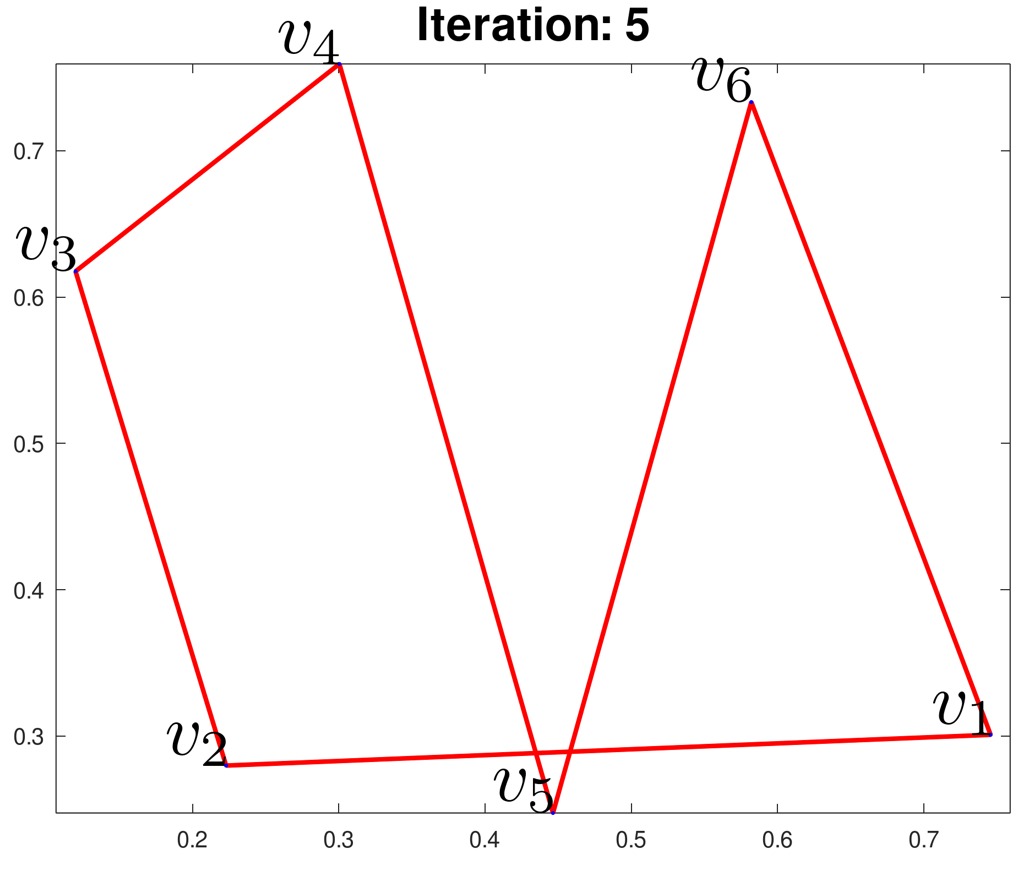}
    \end{minipage} 
    \begin{minipage}{0.32\linewidth}
        \centering
        \includegraphics[width=\linewidth]{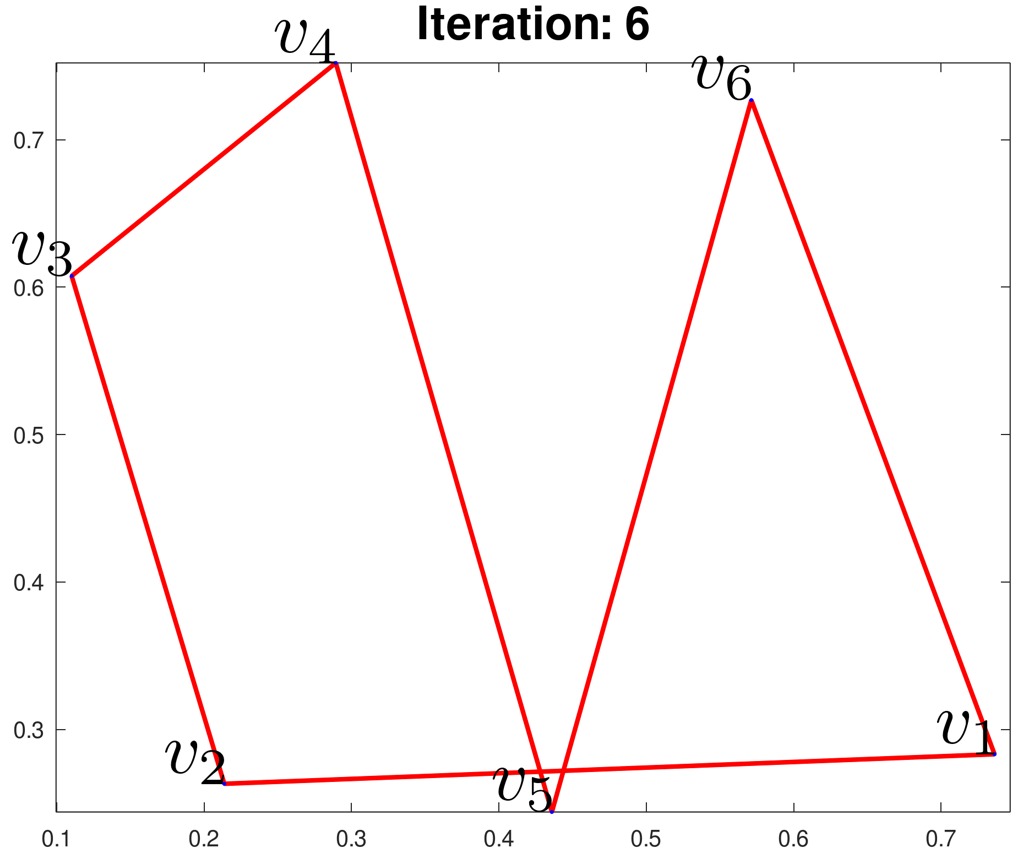}
    \end{minipage}
     \\[0.5em]
    
     \begin{minipage}{0.33\linewidth}
        \centering
        \includegraphics[width=\linewidth]{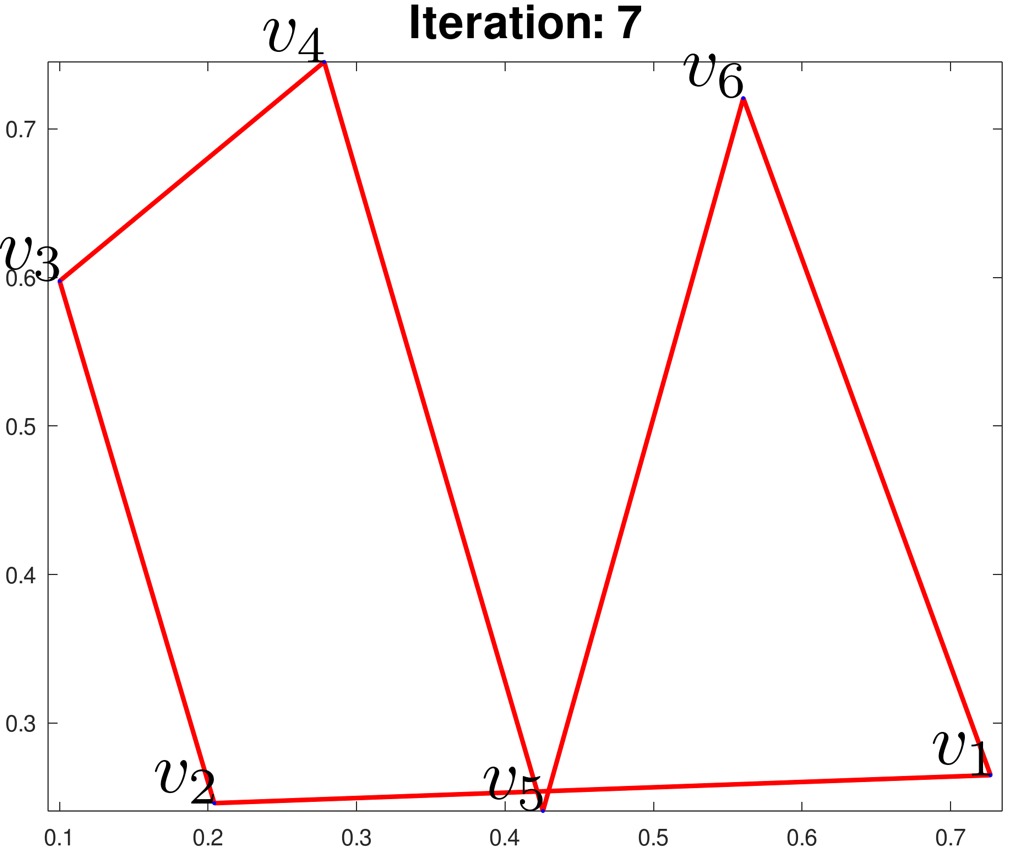}
    \end{minipage}
    \begin{minipage}{0.33\linewidth}
        \centering
        \includegraphics[width=\linewidth]{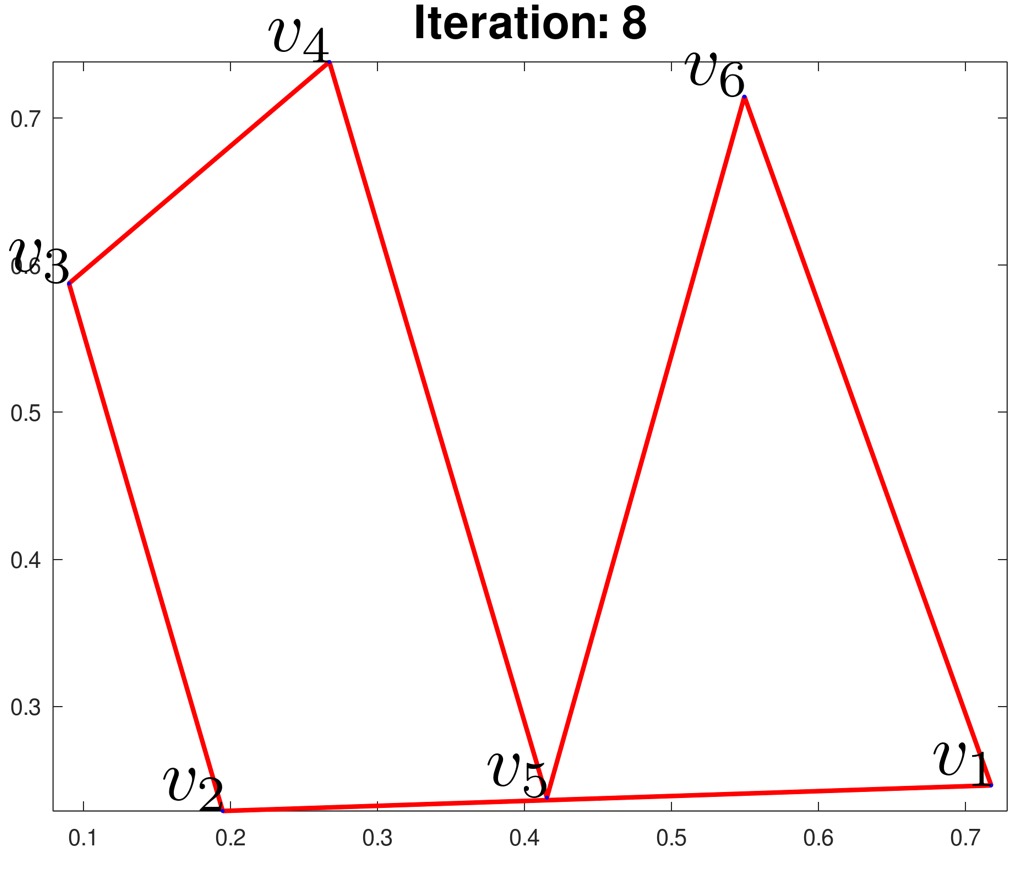}
    \end{minipage} 
    \begin{minipage}{0.32\linewidth}
        \centering
        \includegraphics[width=\linewidth]{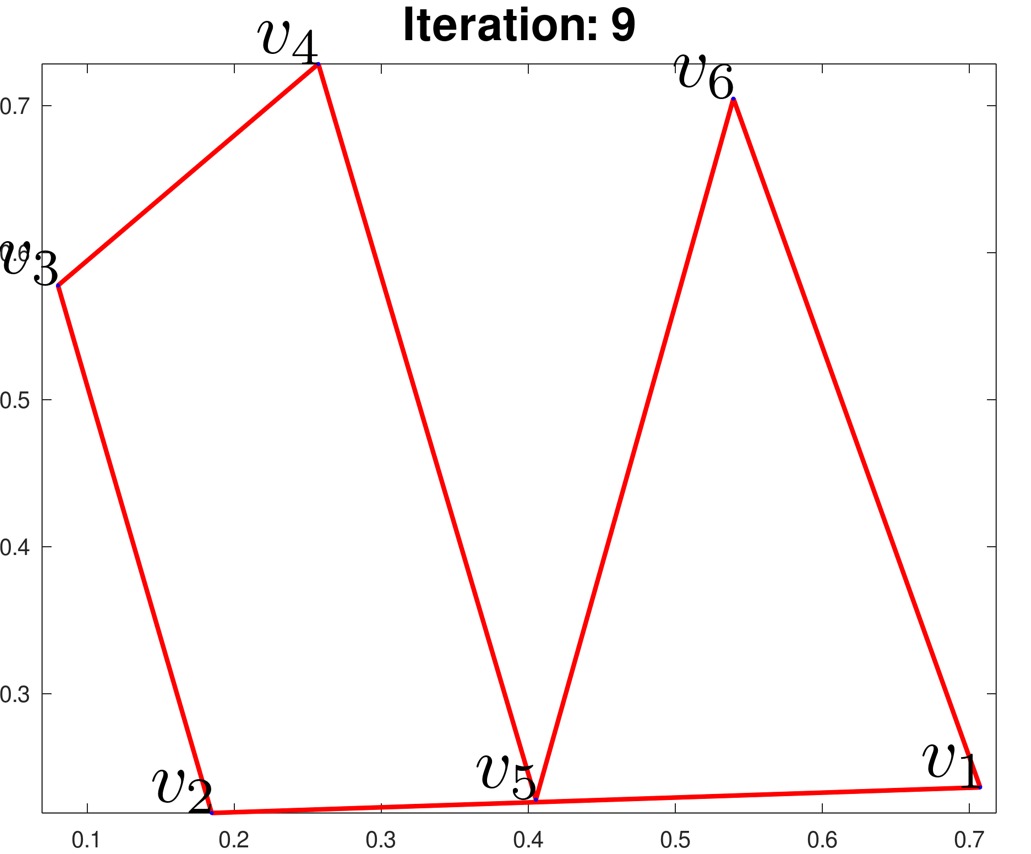}
    \end{minipage}
     \\[0.5em]
    
  \centering
    \begin{minipage}{0.44\linewidth}
        \centering
        \includegraphics[width=\linewidth]{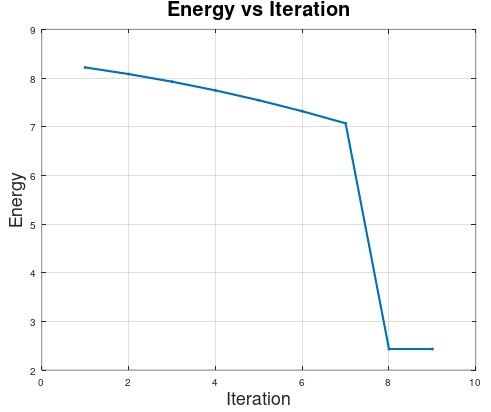}
    \end{minipage}

\caption{Evolution of boundary vertices under the proposed repulsive energy $E_R$. The repulsive force acts on vertex $v_{5}$, progressively resolving the self-intersection. The corresponding energy decay is shown in the bottom plot}
\label{Repulsive_with single contur}
\end{figure}

\begin{figure}[H]
    \centering
    \begin{minipage}{0.42\linewidth}
        \centering
        \includegraphics[width=\linewidth]{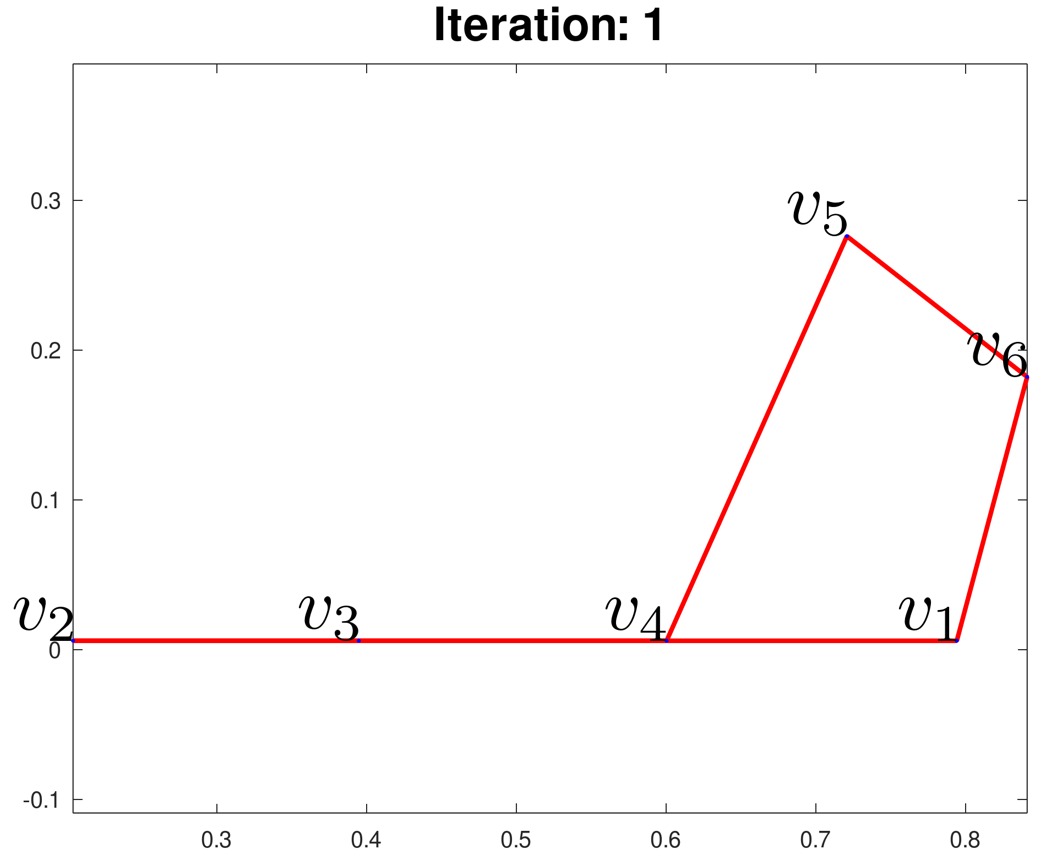}
    \end{minipage}
    \begin{minipage}{0.42\linewidth}
        \centering
        \includegraphics[width=\linewidth]{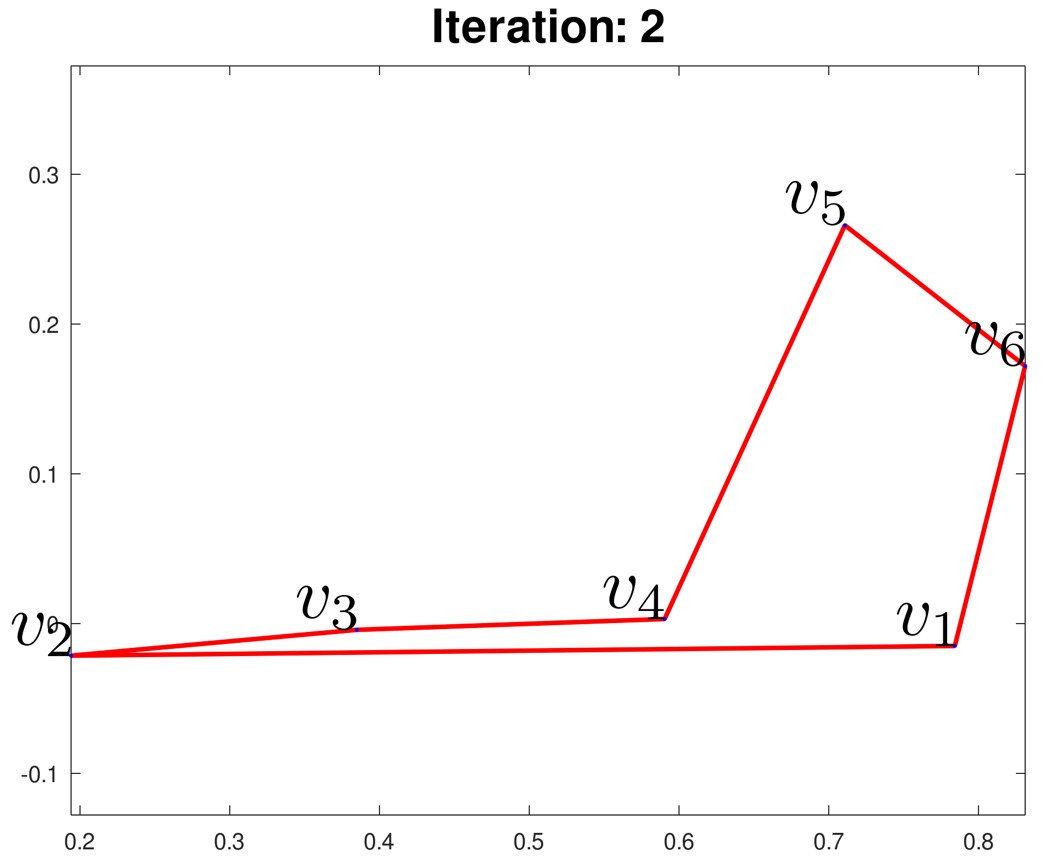}
    \end{minipage}\\
    
    \centering
    \begin{minipage}{0.44\linewidth}
        \centering
        \includegraphics[width=\linewidth]{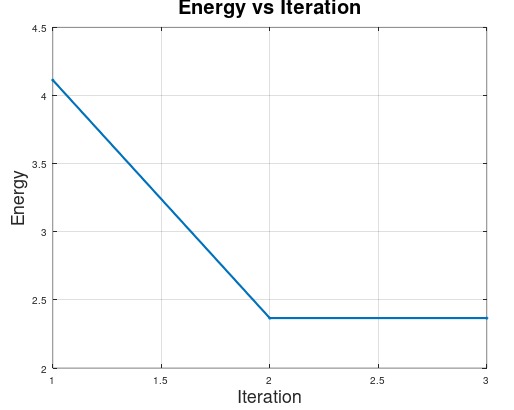}
    \end{minipage}
   
    \caption{Evolution of the boundary when the angular repulsive energy $E_{\theta}$ is activated. The force acting on vertex $v_5$ drives the configuration away from collinearity, accompanied by a steady decrease in the total energy}
    \label{repuslsive with theta term}
\end{figure}

\begin{figure}[h!]
    \centering
    % First Row
    \begin{minipage}{0.34\linewidth}
        \centering
        \includegraphics[width=\linewidth]{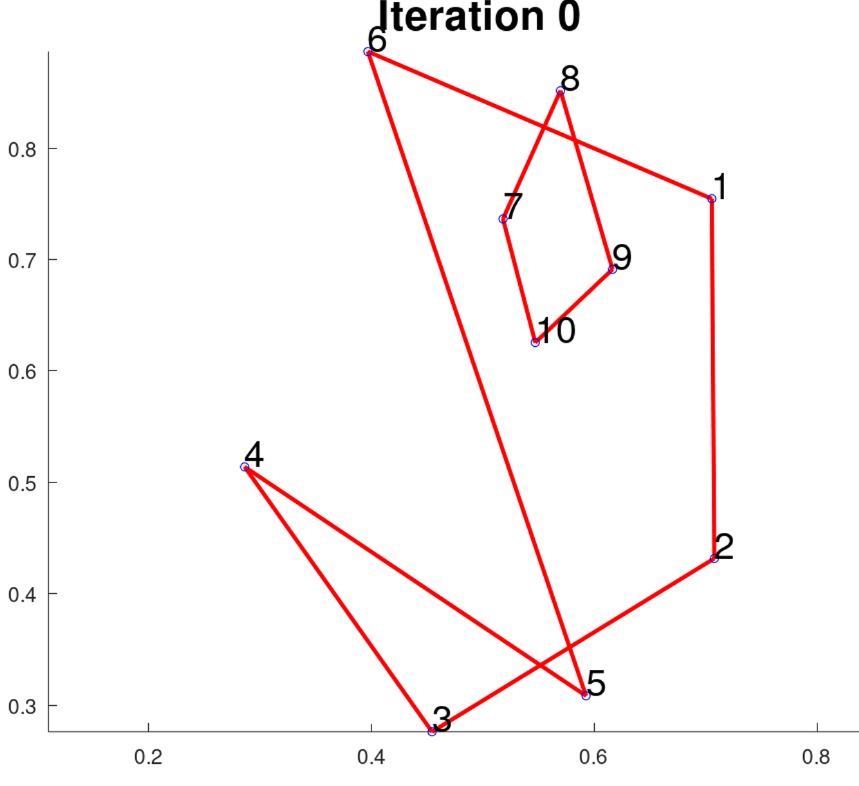}
    \end{minipage}
    \begin{minipage}{0.32\linewidth}
        \centering
        \includegraphics[width=\linewidth]{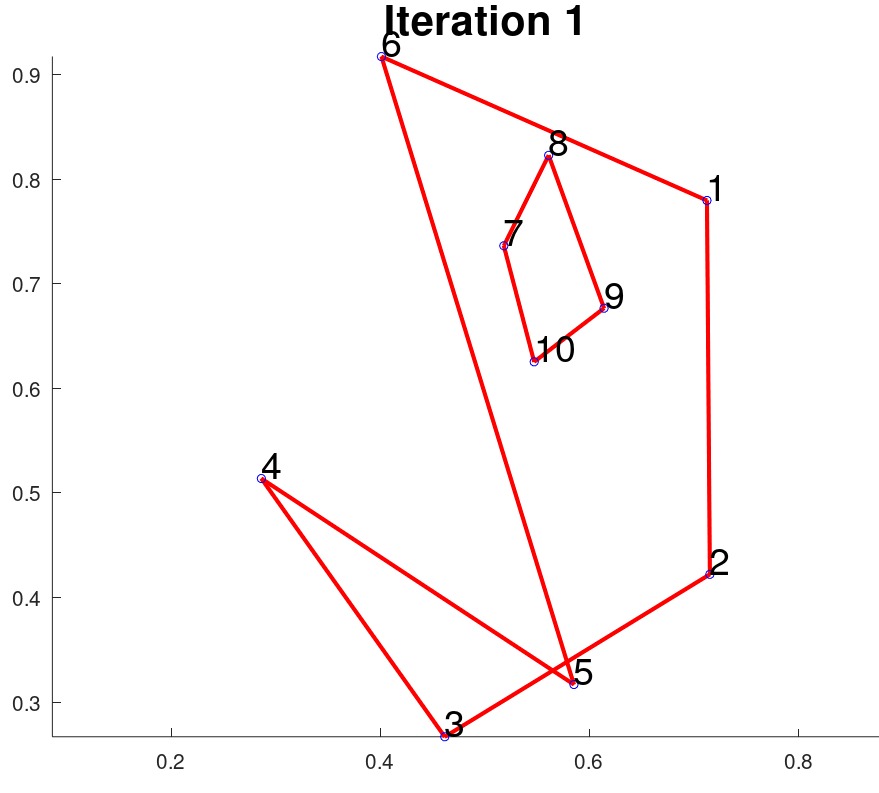}
    \end{minipage} 
    \begin{minipage}{0.32\linewidth}
        \centering
        \includegraphics[width=\linewidth]{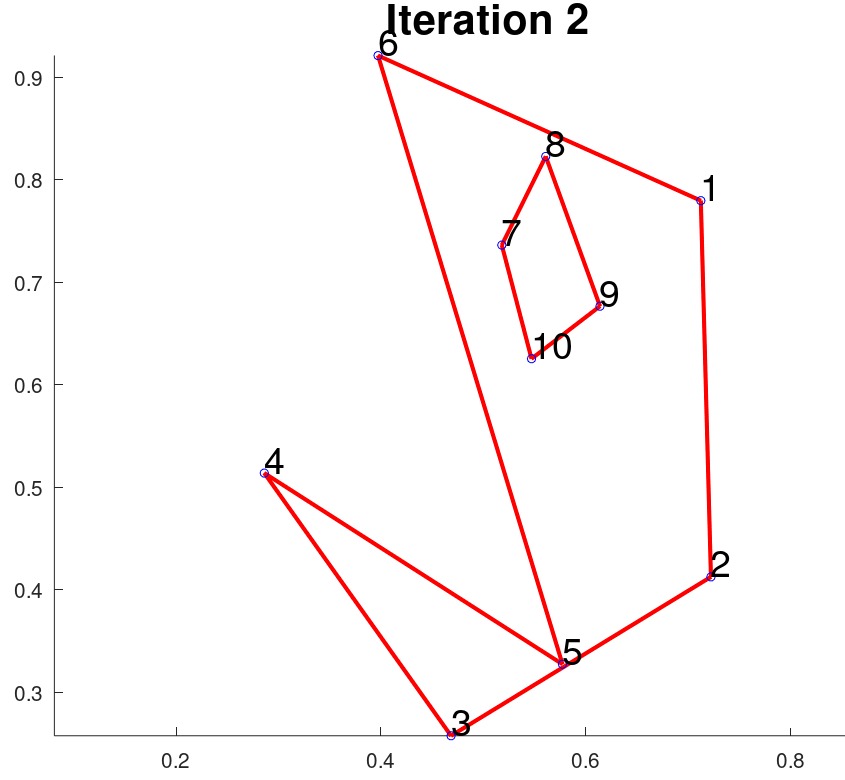}
    \end{minipage}  \\[0.5em]

    \begin{minipage}{0.31\linewidth}
        \centering
        \includegraphics[width=\linewidth]{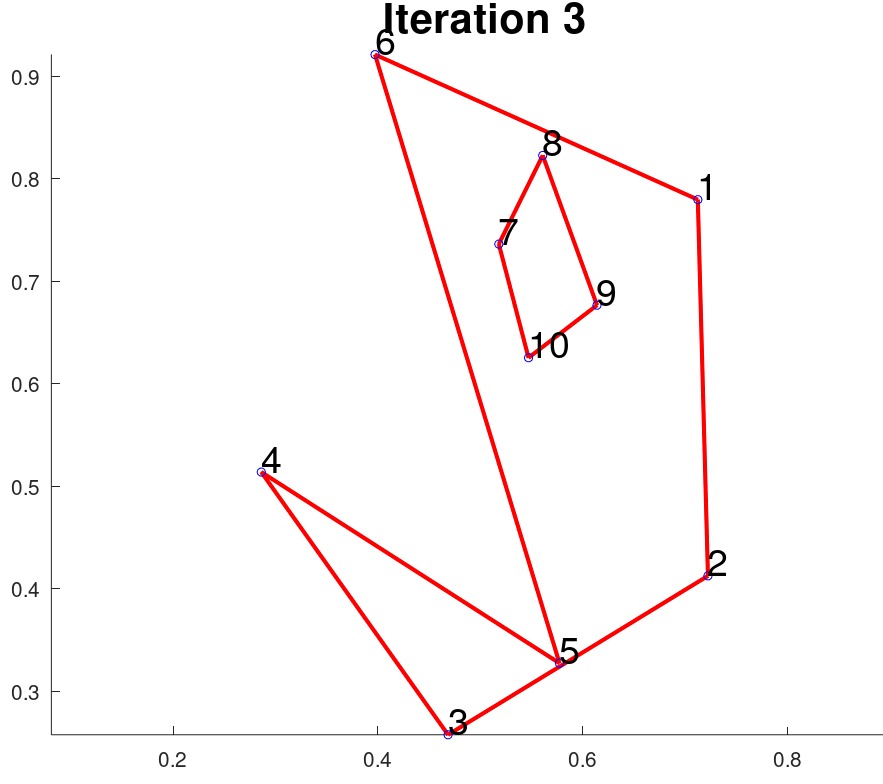}
    \end{minipage}
    \begin{minipage}{0.31\linewidth}
        \centering
        \includegraphics[width=\linewidth]{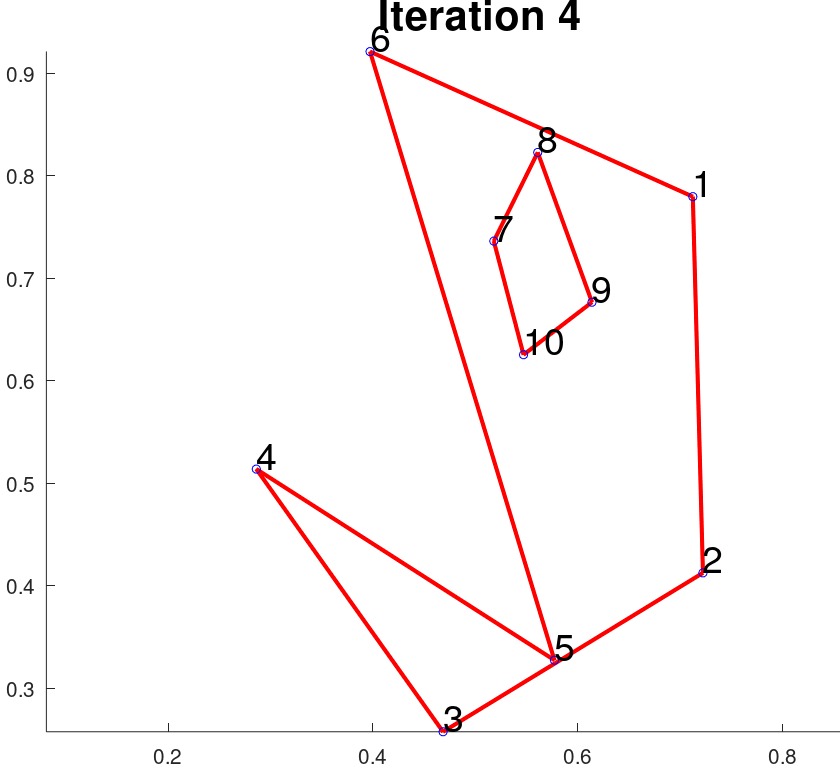}
    \end{minipage} 
    \begin{minipage}{0.35\linewidth}
        \centering
        \includegraphics[width=\linewidth]{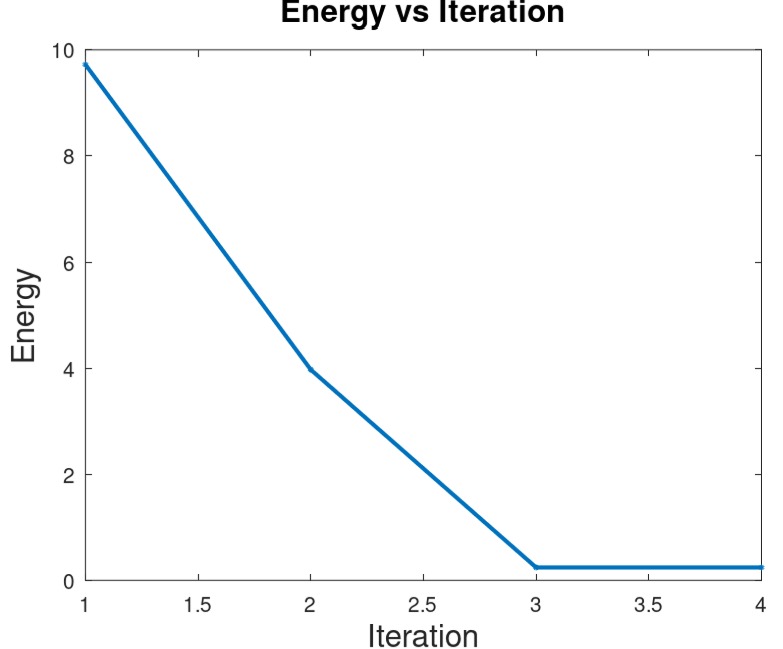}
    \end{minipage}
     \\[0.5em]

\caption{Evolution of boundary vertices under the proposed repulsive energy. The repulsive force acts on vertex $v_{5}$, progressively eliminating self-intersection. The corresponding energy decay is shown in the bottom right plot}
\label{Repulsive_with two contur}
\end{figure}

To analyze the behavior of these repulsive forces, experiments were performed
using both single closed boundary and multiple interacting boundary as initial
configurations. The results are illustrated in
Figures~\ref{Repulsive_with single contur},
\ref{repuslsive with theta term}, and
\ref{Repulsive_with two contur}, where the boundary evolution and corresponding
energy decay are shown over successive iterations.

Figure~\ref{Repulsive_with single contur} demonstrates the action of the
repulsive energy on a single boundary containing a self-intersection.
In this configuration, the intersection occurs between the edge $v_1v_2$ with
the edges $v_4v_5$ and $v_6v_5$, resulting in a unique intersection point in each case.
Consequently, the repulsive term $E_s$ is activated, while the angle-based term
$E_{\theta}$ remains inactive since no collinear intersections are present.
During the evolution, the vertex $v_5$ is gradually displaced inward under the
influence of the repulsive force, and the self-intersection is resolved within a
few iterations. The associated energy plot confirms a monotonic decrease in the
total energy, indicating stable convergence.

The effectiveness of the angle-based repulsive term $E_{\theta}$ is illustrated
in Figure~\ref{repuslsive with theta term}. In this experiment, the initial
boundary contains a degenerate configuration in which the edges $v_1v_2$, $v_2 v_3$ and
$v_3v_4$ are collinear.
In this case, $E_{\theta}$ becomes active while $E_s$ remains zero. The repulsive
force generated by $E_{\theta}$ rapidly perturbs the collinear configuration,
and the degeneracy is resolved within a single iteration and without further side-effects of creating new self-intersections as $E_s$ is also active at the same time as reflected by the sharp drop in the energy curve.

Figure~\ref{Repulsive_with two contur} illustrates the performance of the
repulsive energy in a multi-boundary setting, where two initially disconnected
boundary curves intersect each other. The proposed repulsive mechanism
successfully prevents not only self-intersections within individual boundary
but also intersections between distinct boundary. Within four iterations, the intersecting curves separate, and the repulsive energy stabilizes, leading to a valid and intersection-free configuration.

These experiments demonstrate that the proposed repulsive energy effectively
handles a wide range of intersection scenarios, including unique-point
intersections, collinear degeneracies, and interactions between multiple
contours. The consistent reduction in energy across all cases confirms the
stability and robustness of the proposed framework.

We combine the proposed repulsive energy with the modified Mumford-Shah shape-based
segmentation model, where the evolution of the initial boundary is
driven by a hybrid gradient made of the shape gradient of the Mumford--Shah energy together with the gradient of the repulsive energy.\\
 
Were conducted on a variety of images, including scenes with multiple disjoint objects, images containing interior boundaries, astronomical images with multiple galaxies, color microscopy images
of algae, and real medical images such as CT and MRI scans of bone and brain structures.
In addition, experiments were performed both with and without the repulsive
energy in order to evaluate its influence on segmentation performance. The
results clearly indicate that the repulsive energy plays a crucial role in
obtaining accurate segmentations, particularly in scenarios where object
boundaries are narrow, weak, or closely spaced.

\begin{figure}[h!]
    \centering
    % First Row
    \begin{minipage}{0.40\linewidth}
        \centering
        \includegraphics[width=\linewidth]{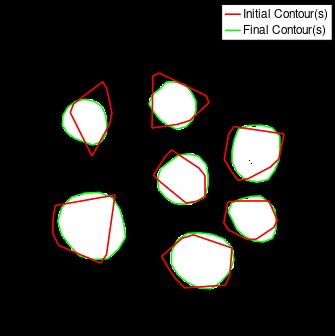}
    \end{minipage}\ \ \ \ \ \ \ \ \ \ \ \  
    \begin{minipage}{0.40\linewidth}
        \centering
        \includegraphics[width=\linewidth]{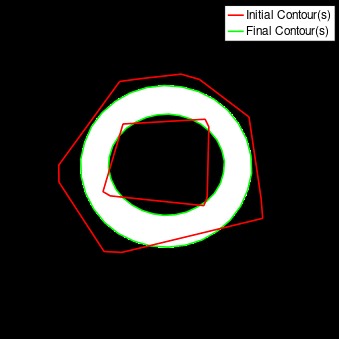}
    \end{minipage} 
   \\[0.5em]
   
   \caption{Segmentation results using multiple boundaries. The red curves denote the initial boundaries, while the green curves represent the final converged boundaries. Left: segmentation of multiple disjoint objects. Right: segmentation of an object with an interior boundary (hole).}
\label{distict object}
\end{figure}
\begin{figure}[h!]
    \centering
    % First Row
    \begin{minipage}{0.40\linewidth}
        \centering
        \includegraphics[width=\linewidth]{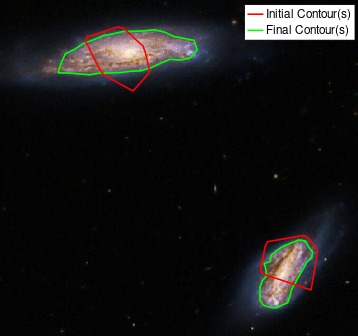}
    \end{minipage}\ \ \ \ \ \ \ \ \ \ \ \ 
    \begin{minipage}{0.40\linewidth}
        \centering
        \includegraphics[width=\linewidth]{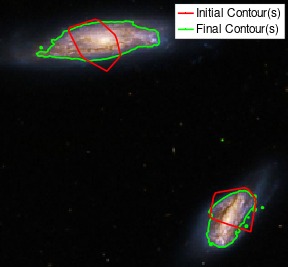}
    \end{minipage} 
   \\[0.5em]

\caption{Comparison of galaxy segmentation results. Left: segmentation obtained using the proposed model. Right: segmentation obtained using the Chan–Vese method \cite{deriche2019color}. Initial boundary are shown in red, and final segmented boundary are shown in green}
\label{Two boundary Image}
\end{figure}

\begin{figure}[h!]
    \centering
    % First Row
    \begin{minipage}{0.32\linewidth}
        \centering
        \includegraphics[width=\linewidth]{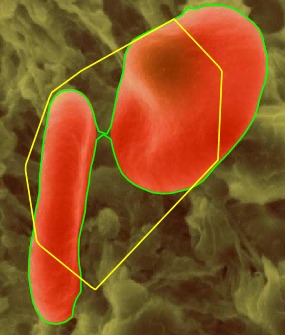}\\ (a)
    \end{minipage}
    \begin{minipage}{0.32\linewidth}
        \centering
        \includegraphics[width=\linewidth]{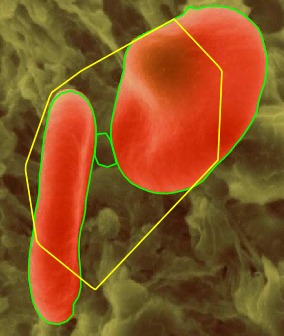}\\ (b)
    \end{minipage} 
    \begin{minipage}{0.32\linewidth}
        \centering
        \includegraphics[width=\linewidth]{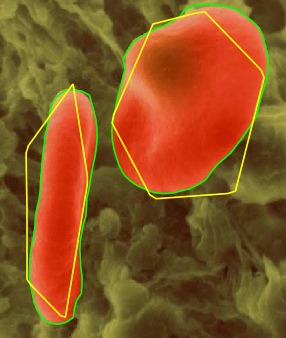}\\(c)
    \end{minipage}  \\[0.5em]

    \begin{minipage}{.60\linewidth}
        \centering
        \includegraphics[width=\linewidth]{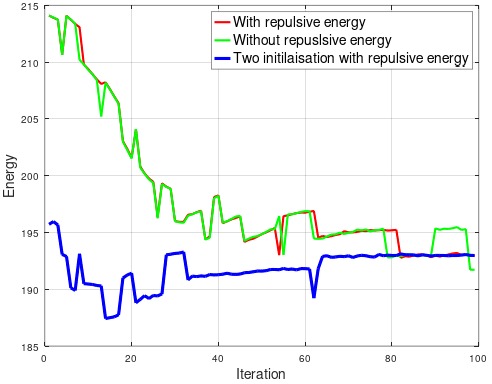}\\(d)
    \end{minipage}
     \\[0.5em]
     \caption{(a) Single boundary initialization with repulsive energy; (b) single boundary initialization without repulsive energy; (c) two boundaries initializations with repulsive energy; (d) corresponding total energy evolution}
\label{alge Image}
\end{figure}

Figure~\ref{alge Image} presents a comparison of image segmentation results
obtained using three different approaches: a simple boundary with repulsion, without repulsion, and multiple boundary initialization. When a single boundary is evolved with the repulsive term, the repulsive interaction becomes active near regions prone to
self-intersection, effectively preventing overlap during evolution. As a
result, the boundary successfully delineates the two algae objects.

In contrast, when the repulsive term is not included, the single evolving
boundary undergoes self-intersection during the evolution process. This
self-intersection leads to incorrect convergence and produces an inaccurate
segmentation. Once self-intersection occurs, the boundary evolution is unable to
recover the correct topology, resulting in an erroneous output.

The third case demonstrates segmentation using two separate initial boundary.
In this configuration, each boundary evolves independently toward a single algae
object, naturally avoiding self-intersections and yielding correct segmentation
of both objects. The corresponding energy plots reflect these behaviors and are
consistent with the observed segmentation results.

%%%%%%%%%%%%%%%%%%%%%%%%%%%%%%%%%%%%%%%%%%%%%%%%%%%%%%%%%%%%%%%%    

%%%%%%%%%%%%%%%%%%%%%%%%%%%%%%%%%%%%%%%%%%%%%%%%%%%%%%%%%%%%%%%%%%%%%%%%%
In the segmentation of medical images, such as CT or MRI scans, evolving boundary are highly susceptible to self-intersection, even when initialized with a single closed curve. This issue mainly arises due to weak image contrast, noise, and the presence of thin or closely spaced anatomical structures, which destabilize the boundary evolution. Such self-intersections lead to invalid curve geometries and often prevent the segmentation process from converging to meaningful anatomical boundaries.

\begin{figure}[h!]
    \centering
    % First Row
    \begin{minipage}{0.40\linewidth}
        \centering
        \includegraphics[width=\linewidth]{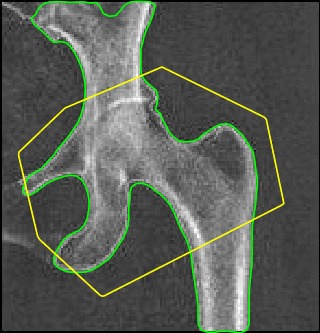}
    \end{minipage}\ \ \ \ \ \ \ \ \ \ \ \ 
    \begin{minipage}{0.40\linewidth}
        \centering
        \includegraphics[width=\linewidth]{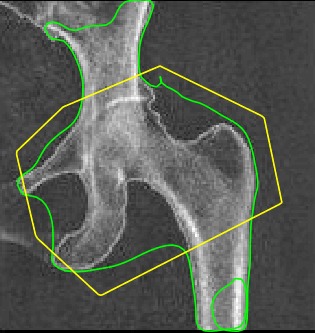
        }
    \end{minipage} 
   \\[0.5em]

\caption{Segmentation results of a CT bone image starting from the same single initial contour, with repulsive energy (left) and without repulsive energy (right).}
\label{Bone segmentation}
\end{figure}

Figure~\ref{Bone segmentation} illustrates the segmentation results of a CT bone image obtained from the same single initial contour, both with and without the proposed repulsive energy term. In the absence of the repulsive energy, the boundary evolution becomes unstable after a few iterations, leading to self-intersections and ultimately resulting in an incorrect segmentation. In contrast, the inclusion of the repulsive energy effectively suppresses self-intersections, preserves the geometric integrity of the evolving boundary, and ensures stable convergence toward the desired anatomical boundary, as shown in Figure~\ref{Bone segmentation}(a).

\begin{figure}[h!]
    \centering
    % First Row
    \begin{minipage}{0.40\linewidth}
        \centering
        \includegraphics[width=\linewidth]{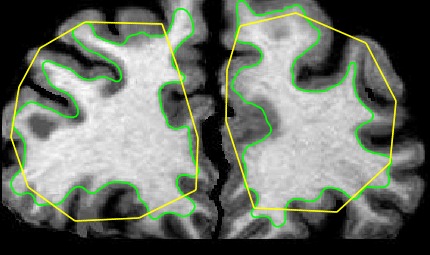}\\(a)
       
    \end{minipage}\ 
    \begin{minipage}{0.40\linewidth}
        \centering
        \includegraphics[width=\linewidth]{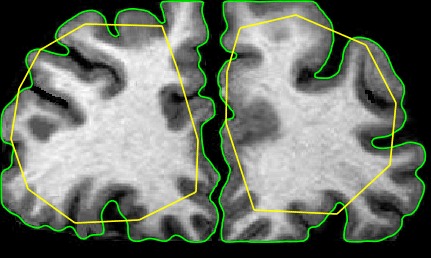
        }\\(b)
    \end{minipage} \\[0.5em]
    \begin{minipage}{0.40\linewidth}
        \centering
        \includegraphics[width=\linewidth]{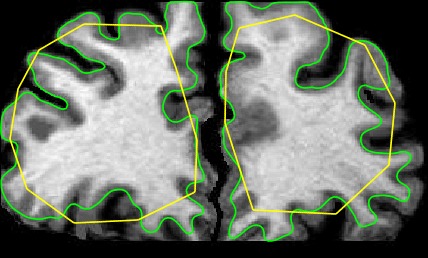
        }\\  (c)
    \end{minipage} 
   \\[0.5em]
    
   \caption{Effect of the parameter ratio $\alpha/\beta$ on brain image segmentation. 
The resulting boundaries are shown for (a) $\alpha/\beta < 1$, 
(b) $\alpha/\beta > 1$, and 
(c) $\alpha/\beta = 1$.}

\label{Image segmentation of MRI}
\end{figure}

Figure~\ref{Image segmentation of MRI} presents the effect of the parameter ratio $\alpha/\beta$ on brain image segmentation in the axial view, where $\alpha$ and $\beta$ denote the weighting coefficients of the foreground and background regions, respectively. The segmentation process is initialized using two closed boundary placed in the left and right hemispheres of the brain, each sampled with 200 vertices. During the evolution, the repulsive energy term prevents interaction between the boundary, allowing them to evolve independently without any intersection. As a result, the evolving curves accurately capture the sulcal structures in each hemisphere.
When $\alpha/\beta < 1$, the background term dominates the evolution, producing smoother boundary that may lack sufficient adherence to anatomical boundaries. Conversely, for $\alpha/\beta > 1$, the foreground term becomes dominant, leading to boundary that more closely follow fine structural details. A balanced weighting, $\alpha/\beta = 1$, provides an optimal trade-off between foreground and background information, yielding stable and accurate segmentation results.

Overall, the proposed modified Mumford--Shah shape model, enhanced with a repulsive energy term, provides a hybrid, robust and effective framework for the segmentation of medical images involving multiple interacting boundaries. By preventing curve intersection and preserving topological consistency, the method ensures stable multi-boundary evolution even in challenging imaging conditions. The experimental results demonstrate the capability of the proposed approach to accurately delineate complex anatomical structures, highlighting its potential applicability to a wide range of medical image segmentation problems involving multiple regions and closely spaced boundaries.

\section{Conclusion}
This work presented a shape-gradient based variational framework for image segmentation that extends the piecewise constant Mumford–Shah model through foreground-background normalization and sampled boundary points evolution and repulsive energy control for topological stability. A key contribution of this work is the introduction of a mathematically grounded repulsive energy that prevents both unique-point and collinear intersections among boundary edges. This repulsive mechanism effectively preserves initial boundary topology during evolution and enables stable segmentation in challenging scenarios involving weak image contrast, narrow(highly convex)
anatomical structures, and closely spaced or interacting boundaries. The formulation seamlessly fits with the shape-gradient descent framework and remains fully differentiable, ensuring numerical stability and consistent energy minimization. 

The extension to multiple independently evolving boundaries allows the proposed model to manually handle images containing disjoint objects, nested regions, and multiple boundaries within a unified variational setting. Extensive experimental results on gray-scale, color, astronomical, and medical images including CT bone
and MRI brain data demonstrate that the proposed approach accurately captures complex structures while avoiding self-intersections that commonly arise in explicit boundary representations.

Overall, the modified Mumford–Shah shape-based framework with repulsive energy  provides
an effective and flexible solution for multi-boundary image segmentation, combining geometric regularization, contrast-based information, and interaction aware boundary evolution. The results highlight its strong potential for applications in medical imaging and other domains requiring absolute control over the topology of the boundary,  accurate delineation of complex and interacting structures.

\subsection*{Data Availability}
  The authors affirm that the data used for experimenting the model will be made available if needed.

\subsection*{Grammar and Readability Disclosure}
  This document has been reviewed with AI-based tools to check grammar and address readability improvements.

\subsection*{Acknowledgment}
The authors thank SRM University, Andhra Pradesh, for providing the fellowship and laboratory facilities that supported this research. Financial assistance for attending various workshops and accessing journal resources has been instrumental in the continuation of this study. 
The authors also sincerely thank the editor and the anonymous reviewers for their careful evaluation of the manuscript and for their valuable comments and suggestions, which greatly improved the quality of the paper.

\subsection*{Disclosure statement}
On behalf of all authors, the corresponding author declares that there are no conflicts of interest related to this work.
\bibliographystyle{apalike}
\bibliography{references}

\end{document}